# Modelling Legal Contracts as Processes

Aspassia Daskalopulu
*Department of Computing, King's College London
Strand, London WC2R 2LS, U.K.
E-mail: aspassia@dcs.kcl.ac.uk*

**Abstract**

*This paper concentrates on the representation of the legal relations that obtain between parties once they have entered a contractual agreement and their evolution as the agreement progresses through time. Contracts are regarded as process and they are analysed in terms of the obligations that are active at various points during their life span. An informal notation is introduced that summarizes conveniently the states of an agreement as it evolves over time. Such a representation enables us to determine what the status of an agreement is, given an event or a sequence of events that concern the performance of actions by the agents involved. This is useful both in the context of contract drafting (where parties might wish to preview how their agreement might evolve) and in the context of contract performance monitoring (where parties might with to establish what their legal positions are once their agreement is in force). The discussion is based on an example that illustrates some typical patterns of contractual obligations.*

**Keywords:** legal contract representation, transition diagrams, temporally qualified obligations, dynamic system modelling.

## 1. Introduction

This paper reports on work that was carried out within a larger project that was concerned with the representation of legal contracts in order to develop tools to support various kinds of contractual activity, such as the management and administration of existing contracts and the drafting of new ones.

The term 'contract' is commonly used to refer both to a legally binding agreement between (at least) two parties and to the document that records such agreement, where it is required that it be put in writing. This paper is concerned with the representation of the legal relations that obtain between parties and therefore focuses on contracts as agreements rather than as documents. Moreover, the larger project within which this work falls was concerned with complex, long-term bilateral agreements rather than one-off or unilateral business exchanges. The contracts that were primarily examined came from engineering areas; for example some concerned the purchase of natural gas from hydrocarbon field owners or the construction of electrical or mechanical plants in the UK and abroad. Such agreements tend to be long-term with a typical life-span of five to ten years and consequently they tend to make provision for a multitude of interrelated issues, some typical of contracts *in general* (such as billing and payment arrangements, delivery times and quantities and so on), some typical of *engineering* contracts in particular (such as insurance arrangements, assignment and sub-contracting terms and so on) and some typical of the particular *kind* of engineering (natural gas purchase contracts for instance contain many provisions about quality and quality monitoring issues whereas construction contracts contain many provisions about amendments to the original plans for the plant in question). Because of their long life-span and the large costs that may be potentially incurred through an omission or a mistake during contract formation or through an inadvertent violation during contract performance, parties tend to spend a considerable time negotiating them and once they are in force each party monitors its own and its counter-party's actions closely. Representing contracts as evolving collections of interrelated contractual obligations is therefore useful in two contexts: During contract formation it enables a preview of each party's obligations under various scenarios that could emerge once the agreement is in force. During contract performance it enables the establishment of the parties' legal relations at a given point in time.

The paper introduces an informal schematic notation that can be used to summarize the structure of agreements as collections of interrelated obligations. Though formal semantics have not yet been developed for the notation, as it stands it enables us to observe the status of legal relations between parties as a contract evolves over time, as a dynamic system. Dynamic systems raise interesting questions about their safety and liveness properties. These are not addressed here, but the interested reader is referred to [4] and [11] for a detailed account of such issues.

## 2. A Typical two-party agreement

A simple two-party agreement will serve to illustrate the main concepts presented here. Consider the following pizza-ordering exchange where a contract is formed between a purchaser (Peter) and a seller (Susan). Although the example is simple on the surface it exhibits some features that are typical of lengthier, more complex agreements: a certain future behaviour is stipulated for the parties, the obligations they assume are time-bounded, provision is made in case an obligation is not fulfilled (the price is reduced in case of late delivery), a third-party (the anonymous driver) is involved, a certain description is stipulated for the goods that are to be delivered and (dynamic) pricing and payment arrangements are made. The example is in this sense a smaller (in size) version of the kinds of agreements that have been analysed in practice.

> **Peter**: Hi, I would like to order a pizza from your menu please.
> **Susan**: Certainly. What kind of pizza would you like and what size?
> **Peter**: The "Good Earth Vegetarian"* please, but without onions. Large, please.
> **Susan**: Very well, that will be £13.95, cash please. What is the address?
> **Peter:** 12 Hunger Lane. How long will that be?
> **Susan:** It is now 7 pm and we promise to deliver within half an hour. If our driver takes any longer than that, we deduct £1.00 from your bill.
> **Peter:** Ok, thank you.
>
> *The menu description of "Good Earth Vegetarian": mushrooms, onions, red and green peppers, all topped with mozzarella.*

Let us first consider a simplified version of the pizza-ordering agreement between Peter and Susan, where detailed temporal elements are ignored. This simple version of the agreement is that Susan is to deliver a pizza (of specified size, quantity and description) and Peter is to pay the required price in return for such delivery.

Let us introduce some elements of the notation that will be used in our discussion of the simplified pizza-ordering agreement. Let $x:\alpha$ stand for "agent $x$ sees to it that $\alpha$". Expressions of the form $\text{not } x : \alpha$ are intended to be read as "it is not the case that agent $x$ sees to it that $\alpha$". The simplified agreement $C$ between Susan and Peter can be represented in terms of the obligations that are imposed on the parties. These can be represented by expressions of the form $O_x \alpha$, which denote that it is obligatory for agent $x$ to see to it that $\alpha$. Such expressions are not necessarily assumed to be closed under logical consequence. As we will see in a moment, we also need a conditional.

The simplified agreement between Peter and Susan goes as follows, where $\alpha$ stands for "delivering a pizza of the specified size, quantity and description to Peter", $\beta$ stands for "paying the required price to Susan", $s$ stands for Susan and $p$ stands for Peter:

It is obligatory for Susan to see to it that a pizza (of specified size, quantity and description) is delivered to Peter, that is, $O_s \alpha$. If Susan fulfils her obligation, then it is obligatory for Peter to see to it that the required price is paid to Susan, that is, if $s:\alpha$ then $O_p \beta$. If Susan does not fulfil her obligation, that is, if $\text{not } s : \alpha$ then it is not obligatory for Peter to see to it that she gets paid, that is, $\neg O_p \beta$. In this case, other norms may also apply. According to Contract Law (cf. [3]) non-fulfilment of contractual obligations yields different consequences, depending on the status of the violated provision. If a party violates a provision that was deemed as a promissory condition, then the counter-party may claim damages and declare the contract void (thus all of the counter-party's obligations under the contract are discharged). On the other hand, if a party violates a provision that was deemed as a warranty, then the counter-party may claim damages, but the contract persists, that is, the counter-party remains bound to his own obligations under the contract. If Susan's obligation $O_s \alpha$ is a promissory condition, then following its breach Peter bears no obligations and she may be obliged to pay damages. That is, where $\delta$ denotes paying damages, if $\text{not } s : \alpha$, then $\neg O_p \beta \land O_s \delta$. If Susan's obligation is a warranty, then following its breach Peter is still obliged to see to it that she gets paid and she is obliged to see to it that damages are paid to Peter, that is, if $\text{not } s : \alpha$, then $O_p \beta \land O_s \delta$. It might be the case that the contract does not specify anything in particular for the case where Susan violates her obligation (perhaps because when the agreement was formed, the status of this obligation as a promissory condition or a warranty was not established, that is, it was left as an intermediate term). In this case following $\text{not } s : \alpha$ the agreement is perhaps terminated and no specific remedies are available to Peter, although possible litigation might ensue to establish whether he might claim damages.

Given these possibilities for the case where Susan violates her obligation, let us assume that the first is the intended one, that is, $\neg O_p \beta \land O_s \delta$ after $\text{not } s : \alpha$. A natural question to pose is what happens if Susan fulfils her obligation to see to it that damages are paid to Peter, that is, what happens after $s : \delta$? One possibility is that the contract returns to its normal course, that is, Peter's original obligation becomes active again ($O_p \beta$). Another possibility is that the contract is terminated. A third possibility is that the contract continues but with different obligations assumed by the parties, for instance $O_s \gamma$ for Susan and some state of affairs $\gamma$, and $O_p \eta$ for Peter and some state of affairs $\eta$. A similar line of questioning can be pursued in relation to Peter's obligation to see to it that Susan gets paid, that is, we can explore what consequences are entailed by the fulfilment or the violation of such obligation.

The foregone discussion highlights the need for a notation that summarizes the structure of the contract between Peter and Susan so that we can determine what obligations hold for them given certain facts that concern the performance or otherwise of actions against given obligations.

One possibility is to represent the contract as a state diagram such as the one illustrated in Figure 1, which summarizes some of the discussion of the simplified pizza-ordering example. Dashed lines indicate the options that were discussed earlier, that is, whether after Susan's violation of her obligation ($S_3$), Peter's original obligation comes into force ($S_1$), or whether the contract is terminated ($S_3$ is final) or whether the contract continues with new obligations for both parties ($S_5$).

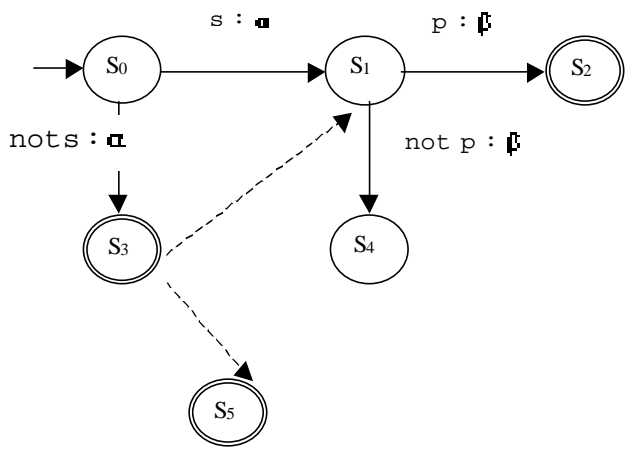

Figure 1 State diagram for simplified pizza-ordering example

Each state offers a (possibly partial) view of the status of the agreement as it evolves given the parties' actions with respect to their obligations. That is, each state corresponds to the obligations (and more generally other legal relations such as rights, powers and so on[1]) that obtain between the parties. Transitions correspond to the performance or non-performance of actions by parties that effect transformations to the status of the agreement. There is no attempt here to provide a formal account of action negation. Expressions such as `not s : α` are merely labels for a kind of transition.

In the state diagram of Figure 1, state $S_0$ corresponds to Susan's obligation $O_s α$ and Peter's obligation $O_p β$ (which is conditional on the fulfilment of $O_s α$). State $S_1$, which follows Susan's fulfilment of her obligation, corresponds to Peter's obligation coming into force. Similarly, state $S_3$, which corresponds to Susan's violation of her obligation, corresponds to $\neg O_p β \wedge O_s φ$ as was noted in the earlier discussion.

## 3. An informal schematic notation

Instead of drawing state diagrams explicitly, we can adapt a notation found in modal languages for transition systems such as modal action logic or dynamic logic.

Expressions of the form $[τ]A$, where $τ$ is the label of a transition, denote that `A` necessarily holds in all states that are reachable following transition $τ$. Expressions such as $\langle τ \rangle A$ are defined as the dual of $[τ]A$, that is, $\langle τ \rangle A =_{def} \neg [τ] \neg A$ ("`A` possibly holds in states that are reachable following transition $τ$"). Therefore, $\langle τ \rangle \mathbf{T}$[2] holds at a state S when there is a transition labelled $τ$ out of S. $[\ ]A$ denotes that `A` is true in all states. This notation appears to be similar to Meyer's deontic logic [8], which is based on dynamic logic. In Meyer's account however, obligations apply to actions. The $τ$ in expressions of the form $\hat{O}τ$ [3] is an action. In the account presented here, obligations apply to states of affairs `Y` in expressions of the form $O_x Y$ and $τ$ is a transition between states.

One state is selected as the initial state and `initially A` means that `A` holds in the initial state. The simplified contract between Peter and Susan can now be represented as follows, where `terminated` is a special symbol denoting a final state for the agreement:

$$\text{initially} \begin{cases} O_s α \\ \langle s : α \rangle \mathbf{T} \\ \langle \text{not } s : α \rangle \mathbf{T} \end{cases}$$

$[\ ](O_s α \to [s : α]O_p β)$
$[\ ](O_s α \to [\text{not } s : α]\neg O_p β \wedge O_s φ))$
$[\ ](O_s φ \to [s : φ]O_p β)$
$[\ ](O_s φ \to [\text{not } s : φ]\text{terminated})$
$[\ ](O_p β \to [p : β]\text{terminated})$
$[\ ](O_p β \to [\text{not } p : β]\text{terminated})$

Rules such as the above implicitly define the state space for the agreement. Initially it is obligatory for Susan to see to it that a pizza gets delivered to Peter. It is possible for Susan to see to it that a pizza gets delivered to Peter and it is also possible for her not to see to it that this state of affairs is achieved. A transition labelled `s : α` from any state in which Susan's obligation holds necessarily leads to a state where it is obligatory for Peter to see to it that Susan gets paid. A transition labelled `not s : α` from any state where Susan's obligation is in force necessarily leads to a state in which Peter is not obliged to see to it that she gets paid and Susan is obliged to see to it that Peter receives damages. Peter's obligation to see to it that Susan gets paid becomes active in all states following either fulfilment of $O_s α$ or fulfilment of $O_s φ$. The agreement is terminated 'happily' following Peter's fulfilment of his obligation or 'unhappily' (with possible litigation ensuing) following Susan's failure to see to it that damages are awarded to Peter (after she fails to deliver the pizza) or Peter's failure to pay the required price after Susan successfully delivers the pizza.

---

[1] Hohfeld's account [5] of legal notions such as obligation, right, power, immunity and so on and their interrelationships has been adopted.

[2] $\mathbf{T}$ is the constant symbol for "true".

[3] $\hat{O}$ is Meyer's obligation operator.

In addition, the following rules are built in, where `x` is a variable ranging over agents' labels and `Y` is a variable ranging over propositional states of affairs:

(i) $[\ ]O_xY \rightarrow \langle x : Y\rangle T)$ that is, where an agent bears an obligation it is possible for him to perform an action that fulfils it;

(ii) $[\ ]Pow_xY \rightarrow \langle \Pi_xY\rangle T)$ that is, where an agent has power[4] to effect a legal relation `Y` it is possible for him to exercise such power;

(iii) $[\ ]Pow_xY \rightarrow [\Pi_xY]Y)$ that is, where an agent has power to effect a legal relation `Y`, the legal relation obtains in all states following his exercise of such power[5].

We could also consider building in the following:

$[\ ]O_xY \rightarrow \langle \text{not } x : Y\rangle T)$

That is, where an agent bears an obligation it is possible for him to violate it. As we shall see later however, this is a point worthy of more discussion.

In the representation of the simple pizza-ordering agreement above, we observe structures that are similar to contrary-to-duty obligations. For example, concentrating on Susan's obligation we have the following[6]:

$[\ ]O_s\alpha \rightarrow [s : \alpha]O_p\beta)$

$[\ ]O_s\alpha \rightarrow [\text{not } s : \alpha]\neg O_p\beta)$

$[\ ]O_s\alpha \rightarrow [\text{not } s : \alpha]O_s\phi)$

That is, if Susan violates her primary obligation $O_s\alpha$ then a secondary obligation $O_s\phi$ comes into effect. Representing and reasoning about contrary-to-duty structures in a consistent way has been the topic of much debate (cf. [1]). The consensus seems, however, to be that in temporal settings contrary-to-duty structures are not problematic (cf. [9, 10]).

At first glance, it might appear that contrary-to-duty structures are a common feature of contracts. They seem to emerge from contractual provisions that specify reparations in cases where an obligation is violated. Usually such provisions are associated with warranties. Contractual provisions are not typically explicitly labelled as warranties, promissory conditions or intermediate terms. Normative propositions that stipulate secondary obligations can be used to establish whether normative propositions specifying primary obligations are warranties. In contracts, it is more common for the violation of a primary obligation by one party `x` to be associated with power for the counter-party `z` to bring about a legal relation `R`. That is, in general, the following pattern is typically encountered:

$[\ ]O_xY \rightarrow [\text{not } x : Y]Pow_zR)$

Such legal relation `R` might entail that the contract is declared void (terminated) if the primary obligation that was violated is treated as a promissory condition. Alternatively, such legal relation `R` might entail new obligations for the offending party. For simplicity, in the pizza-ordering example Susan's violation of her primary obligation entailed a new obligation for her (to see to it that damages are paid to Peter $O_s\phi$). It is more accurate to say that Susan's violation of her primary obligation entails Peter's power to create a new obligation $O_s\phi$ for her. That is:

$[\ ]O_s\alpha \rightarrow [\text{not } s : \alpha]Pow_pO_sF)$

Whether this new obligation will actually come into force depends on whether Peter chooses to exercise his power.

In discussing violation of obligations, we are interested in transitions with negative labels. Let us revisit the original pizza-ordering example, in which Susan's obligation to see to it that a pizza is delivered to Peter was time-bounded and the pizza in question had to conform to some requirements of size, quantity and description. The transition `not s : `$\alpha$ denotes that Susan violates her obligation. Such violation however may come about in different ways. For example, Susan might not deliver a pizza at all. Or a pizza might be delivered, but it might not conform to the agreed specification (in terms of size, quantity and description), that is, Susan brings about something which is different from what she is obliged to bring about (`s : `$\alpha'$). A third possibility is that the 'right' (in terms of size, quantity and description) pizza is delivered but not within the agreed time bounds; again Susan brings about something which is different from what she is obliged to bring about (`s : `$\alpha''$). A fourth possibility is that the 'wrong' pizza gets delivered late (`s : `$\alpha'''$). Other possibilities exist by considering various combinations of what might potentially count as violation of Susan's primary obligation (for instance, the 'right' pizza is delivered on time but not by Susan, the 'wrong' pizza is delivered on time by somebody other than Susan, or the 'wrong' pizza is delivered late and by somebody other than Susan and so on). Instead of a single transition labelled `not s : `$\alpha$, emerging from a state in which $O_s\alpha$ holds, we have a cluster of refined transitions, each corresponding to one aspect of "it is not the case that Susan sees to it that a pizza of the required size, quantity and description is delivered to Peter within specified time bounds". This is illustrated in Figure 2, where

A ■ ($\alpha \vee \alpha' \vee \alpha'' \vee \alpha'''$):

---

[4] "Power" here refers to legal power or legal competence (cf. [5]).

[5] This is a simplifying assumption for the purposes of this discussion. Bearing legal power does not generally entail having the ability to exercise it (cf. [4] for further discussion).

[6] Transition labels distribute over compound propositional formulae.

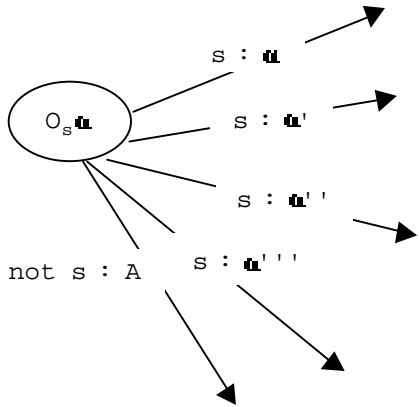

Figure 2 Possible transitions following Susan's obligation

It should be stressed again that these are merely labels for transitions and no attempt is made to provide a formal account of negation in action descriptions. Moreover, as Figure 2 suggests, concurrent actions are in principle possible from a state but no attempt is made here to address the logic of concurrent action.

Exploring the consequences entailed by the violation of an obligation raises other interesting points. For example, as was mentioned above, there are various possibilities, which can be construed as violations of Susan's obligation towards Peter. One way of perceiving the agreement between Susan and Peter is as follows: Susan is obliged to deliver a pizza of any type from a range of available types[7], say *t1* or *t2*. If Susan delivers a pizza of type *t1,* then Peter is obliged to pay a price *p1*. If Susan delivers a pizza of type *t2,* then Peter is obliged to pay a price *p2*. Another way of perceiving the agreement between Susan and Peter is this: Susan is obliged to deliver a pizza of type *t1*. If she delivers a pizza of type *t1,* then Peter is obliged to pay a price *p1*. If she delivers a pizza of type *t2,* then Peter is obliged to pay a price *p2* and Susan is no longer obliged to deliver a pizza of type *t1*. The difference between the two versions is subtle. In the first version, Susan is only obliged to deliver *a* pizza. The exact amount that Peter is obliged to pay in return for such delivery is dependent on what particular kind of pizza is delivered. Nothing seems to preclude the provision of a pizza of type *t1* after a pizza of type *t2* has been successfully delivered. In the second version, Susan is obliged to deliver a *specific* pizza. Non-fulfilment of her obligation to deliver a pizza of type *t1* discharges such obligation and possibly leads to secondary (reparation) obligations on her part that are effected through Peter's exercise of power. The gas-purchase sample contracts that we examined tend to be like the second version. They specify precisely what happens, if the natural gas that is delivered does not conform to pre-agreed quality standards, or if delivery quantities and times depart from those stipulated in the contract (for instance prices are adjusted, additional quantities must be delivered at more frequent intervals to make up for the under-delivery and so on).

Let us now turn to the temporal elements of the original pizza-ordering example. Our informal notation for representing agreements implicitly defines a state space. In principle, time can be associated with states (which correspond to the collection of obligations and other legal relations between parties) or with transitions (which correspond to actions performed by parties). That is, expressions of the form $O_x^t Y$ can be used to denote that at time $t$, agent $x$ is obliged to see to it that $Y$ is the case. On the other hand, expressions of the form $O_x Y^t$ can be used to denote that agent $x$ is obliged to see to it that $Y$ is the case at time $t$. In the latter form, time is implicitly associated with the agent's action. Put alternatively, in the latter case time is associated with the transition $x : Y$. We show this explicitly in our informal notation by extending the labels for transitions to include time in expressions of the form $(x : Y)^t$, in which time is specified in absolute or relative terms, for example via temporal relations such as before(t), after(t), or between(t1, t2). Let p denote "paying a reduced price to Susan", π denote "a pizza of the specified size, quantity and description is delivered to Peter" and β denote "paying the normal price to Susan". The state diagram for the pizza-ordering example is now as illustrated in Figure 3:

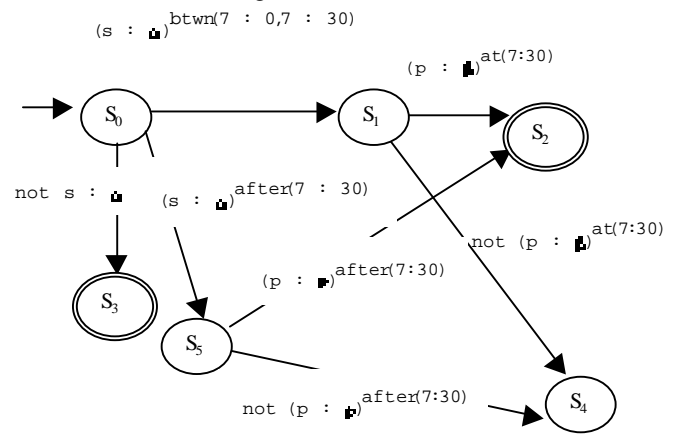

Figure 3 State diagram for pizza-ordering example with temporal elements

Note that time is here used informally as a label on transitions—no attempt is made here to define formal semantics for it.

## 4. Concluding Remarks and Future Work

Such representation offers a process view of the contract and enables us to determine what the status of the agreement is (what obligations hold), given an event or a sequence of events that concern the performance of actions by agents. As the underlying model is so familiar in Artificial Intelligence (for example situation calculus [7], or simplified

---

[7] "Type" here is intended to refer to size, quantity, description and possible time restrictions that apply.

event calculus [6]), it would be feasible to build an implementation.

In representations such as this, where events effect transitions between states, the initiation or termination of properties that hold in states (norms in this case) can be expressed through the definition of appropriate relations (such as *initiates* or *terminates*) between events and states. Such relations essentially define the post-conditions of events. As mentioned previously, states in our state diagram do not necessarily offer a complete view of the contract. We have only been showing activated obligations in states that are arrived at through some transition, and we have assumed that any obligations from the previous state that are not explicitly stated are assumed to be discharged. In addition to establishing what norms become activated or discharged, we are also interested in what norms persist over transitions between states. This is one of the subjects of ongoing work in the area of temporally qualified obligations (for example [2] and the discussion in [10]) and has not been dealt with in this work. It is however relevant to possible future extensions to our proposed representation of contracts as processes. As was noted earlier we do not provide a formal account for action negation (negated transition labels) and time constraints here. This is another possible avenue for future work.

**Acknowledgements**: *I gratefully acknowledge the helpful involvement and generous sponsorship of British Gas Plc. during the first years of the broader project (1992–1995). I am particularly indebted to Marek Sergot for numerous ideas and inspiring discussions and to Trevor Bench-Capon and Peter Hammond for their valuable comments.*